\documentclass[letterpaper]{article} 
\usepackage{aaai25}  
\usepackage{times}  
\usepackage{helvet}  
\usepackage{courier}  
\usepackage[hyphens]{url}  
\usepackage{graphicx} 
\usepackage{natbib}  
\usepackage{caption} 
\usepackage{algorithm}
\usepackage{algorithmic}

\usepackage{newfloat}
\usepackage{listings}
\DeclareCaptionStyle{ruled}{labelfont=normalfont,labelsep=colon,strut=off} 
\floatstyle{ruled}
\newfloat{listing}{tb}{lst}{}
\floatname{listing}{Listing}
\usepackage{multirow}
\usepackage{array}
\usepackage{graphicx} 
\usepackage{subcaption} 
\usepackage{amsmath}  
\usepackage{amssymb}  

\usepackage{pgfplots}
\pgfplotsset{compat=1.17}

\title{DMPA: Model Poisoning Attacks on Decentralized Federated Learning for Model Differences}

\author {
    Chao Feng\textsuperscript{\rm 1}, 
    Yunlong Li\textsuperscript{\rm 1}, 
    Yuanzhe Gao\textsuperscript{\rm 1}, 
    Alberto Huertas Celdrán\textsuperscript{\rm 1}, 
    Jan von der Assen\textsuperscript{\rm 1},
    Gérôme Bovet\textsuperscript{\rm 2}, 
    Burkhard Stiller\textsuperscript{\rm 1}
}
\affiliations{
    \textsuperscript{\rm 1} Communication Systems Group, Department of Informatics, University of Zürich,\\
    Binzmühlestrasse 14, CH-8050 Zürich, Switzerland\\
    \textsuperscript{\rm 2} Cyber-Defence Campus, armasuisse Science \& Technology, CH-3602 Thun, Switzerland\\
    \{cfeng, huertas, vonderassen, stiller\}@ifi.uzh.ch, yuanzhe.gao@uzh.ch, yunlong.li@uzh.ch, gerome.bovet@armasuisse.ch
}

\usepackage{bibentry}

\begin{document}

\maketitle

\begin{abstract}
Federated learning (FL) has garnered significant attention as a prominent privacy-preserving Machine Learning (ML) paradigm. Decentralized FL (DFL) eschews traditional FL's centralized server architecture, enhancing the system's robustness and scalability. However, these advantages of DFL also create new vulnerabilities for malicious participants to execute adversarial attacks, especially model poisoning attacks. In model poisoning attacks, malicious participants aim to diminish the performance of benign models by creating and disseminating the compromised model. Existing research on model poisoning attacks has predominantly concentrated on undermining global models within the Centralized FL (CFL) paradigm, while there needs to be more research in DFL. To fill the research gap, this paper proposes an innovative model poisoning attack called DMPA. This attack calculates the differential characteristics of multiple malicious client models and obtains the most effective poisoning strategy, thereby orchestrating a collusive attack by multiple participants. The effectiveness of this attack is validated across multiple datasets, with results indicating that the DMPA approach consistently surpasses existing state-of-the-art FL model poisoning attack strategies.

\end{abstract}

%
\section{Introduction}

The advent of diverse privacy regulations has significantly increased the need for privacy preservation in Machine Learning (ML). Federated Learning (FL) emerges as a novel collaborative and privacy-preserving ML paradigm that has attracted substantial attention from both academic and industrial communities~\cite{mcmahan2017communication}. FL enables participants to share model parameters instead of raw data, thereby ensuring data privacy \cite{yang2019federated}. Traditional FL architectures rely on a central server to distribute, called Centralized FL (CFL), receive and aggregate model parameters from participants \cite{alazab2021federated}. However, this client-server architecture suffers from significant drawbacks, including susceptibility to a single point of failure and server-side bottleneck~\cite{Mart_nez_Beltr_n_2023}. To address these challenges, Decentralized FL (DFL) has been introduced.


In DFL, each client independently manages the processes of establishing communication, sharing, and aggregating model parameters with other clients \cite{lian2017can}. During each iteration, participants execute several tasks: training their local models, transmitting model parameters to directly connected peers, aggregating received parameters, and updating their local models accordingly. Unlike CFL, clients in DFL have the autonomy to independently select any other participant to establish bidirectional communication connections, allowing for customizable overlay network topology, such as fully connected, ring, star, and even dynamic configurations \cite{yuan2024decentralized}. This adaptability enables DFL to effectively address the single point of failure issue inherent in CFL, thereby enhancing the system's robustness and scalability. Nevertheless, this decentralized nature introduces new security vulnerabilities, making DFL more susceptible to various threats, including model poisoning attacks.

Model poisoning attacks involve malicious clients directly altering their local model parameters to negatively influence the global model training by sending harmful updates \cite{feng2023sentinelaggregationfunctionsecure}. In FL, since participants can directly share and modify model parameters, model poisoning attacks are relatively more straightforward to execute and pose significant threats. Current model poisoning attacks primarily target CFL models. These adversarial clients transmit meticulously crafted local model updates to the central server during the training phase, leading to issues such as reduced accuracy of the global model \cite{cao2022mpaf}. These attacks presuppose that the attacker controls a substantial number of compromised real clients, which may include either hijacked clients or fabricated ones.

Nevertheless, the architectural distinctions between DFL and CFL, particularly in overlay network topology, imply that existing attacks targeting CFL can not be directly transferable to DFL. In DFL, the communication links are significantly longer compared to CFL, making it challenging for a malicious participant to impact the entire federation \cite{Mart_nez_Beltr_n_2023}. Additionally, in DFL, each node could decide whether to disseminate and receive models from neighboring nodes. This capability enables the potential blockage of malicious attacks at intermediate nodes, thereby diminishing the effectiveness of such attacks. Consequently, from an adversarial perspective, it is imperative to develop model poisoning attacks that are effective in DFL.

To this end, this paper investigates model poisoning attacks within DFL  and proposes the Decentralized Model Poisoning Attack (DMPA) based on an Angle Bias Vector to rectify reversed model parameters. Specifically, by determining the eigenvalues of the compromised benign parameters, the method extracts the corresponding eigenvectors to compute the angular deviation vector. This vector is derived by exploiting the discrepancies between models and ultimately adjusting the negative parameters, thereby generating an effective attack model across the majority of participants. The main contributions of this paper are: \textit{(i)} design and implementation of a model poisoning attack, called DMPA, to compromise the model robustness of DFL models; \textit{(ii)} an extensive series of experiments are conducted to evaluate the proposed DMPA and compared with selected state-of-the-art attack algorithms under diverse robustness aggregation functions, which encompass three benchmark datasets (MNIST, Fashion-MNIST, and CIFAR-10) and three varying overlay topologies (fully connected, star, and ring); and \textit{(iii)} critical insights into the robustness of the DFL model concerning both offensive and defensive points of view. The experiments demonstrate that the DMPA presented in this work exhibits a more potent attack capability and broader propagation potential, effectively compromising the DFL system.

The remainder of this paper is structured as follows. Section 2 provides an overview of the underlying security issues in DFL and discusses the concept of model poisoning attacks. Subsequently, Section 3 formalizes the problem. Section 4 delineates the design specifics of DMPA, the proposed model poisoning attack. Section 5 then evaluates DMPA's performance in comparison to selected related works. Finally, Section 6 offers a summary of the contributions made by this research and outlines potential avenues for future investigation.

\section{Background and Related Work}
This section provides an introduction to the security problem in DFL and a summary of the current research on model poisoning attacks in FL.
\subsection{Security Problem in DFL}
Differing from traditional CFL, neighboring clients in DFL exchange local model parameters or gradients over a P2P network and construct consensus models independently. DFL solves the problem of risk associated with a single point of failure, enhances its scalability, and is well suited for applications in the Industrial Internet of Things (IIoT) \cite{tan2023collusive}. However, the decentralized nature of DFLs rather increases their risk of being exposed to malicious attacks, especially poisoning attacks~\cite{feng2023voyager}. Poisoning attacks, which aim to diminish the resilience of FL models, can be classified into data poisoning and model poisoning attacks. Data poisoning attacks typically involve malicious clients interfering with the training process by introducing harmful data (such as backdoor attacks or label flipping) into the local training dataset, thereby inducing biased learning outcomes \cite{hallaji2022federated}. Conversely, model poisoning attacks are characterized by malicious clients directly altering their local model parameters and affecting the global model training via harmful model updates. To improve the model robustness of FL and defend against the poisoning attacks, \cite{mcmahan2017communication} proposed the use of averaging model parameters to improve training efficiency. \cite{blanchard2017machine} introduced the Krum method, which excludes malicious updates by calculating the Euclidean distance of a multi-node model and selecting the model with the smallest distance. This method reduces the impact of malicious attacks and is now widely used for robust aggregation in DFL.
\cite{yin2018byzantine} developed two robust distributed learning algorithms for Byzantine errors and potential adversarial behavior, a robust distributed gradient descent algorithm based on Median and Trimmed Mean operations, respectively. These two methods are widely used in both CFL and DFL because they can be implemented with only one communication round and achieve optimal model robustness.

\subsection{Model Poisoning Attack}
Compared with data poisoning attacks, model poisoning attacks are easier to execute as they directly modify the shared model. Therefore, there has been research suggesting how to optimize the attack method to enhance the attack effectiveness. Existing model poisoning attacks mainly target CFL. The attack assumes that the attacker has access to a large number of compromised clients. During the training process, malicious clients send carefully designed local model updates to the server, resulting in problems such as a decline in the accuracy of the global model \cite{cao2022mpaf}. The MIN-MAX and MIN-SUM methods proposed by \cite{shejwalkar2021manipulating} respectively minimize the maximum and sum of the distance between toxic updates and all benign updates in CFL. These methods assume that malicious model updates are the sum of aggregated model updates and a fixed disturbance vector scaled by factors before the attack, and implement the attack by maximizing the distance between toxic updates and benign updates in the inverse direction of the global model. 
The "a little is enough" (LIE) attack \cite{baruch2019little} directly modifies the model parameters, which generates malicious model updates in each training round by calculating the average of the real model updates of the malicious client and perturbing them. 

Most model poisoning attacks rely on additional information from the central server, such as aggregation methods, global models, or even the training data utilized by nodes. This dependency renders external attacks impractical. Nevertheless, some attacks are engineered to maintain efficacy even in the absence of such additional knowledge. For instance, \cite{zhang2023denial} employs global historical data to build an estimator that forecasts the subsequent round of global models as a benign reference. Despite this, the approach necessitates substantial resources and encounters difficulties in accessing the global model within DFL systems, thereby limiting its practical application in DFL.

To conclude, while a substantial amount of research has been dedicated to optimizing model poisoning attacks against the FL, these efforts predominantly concentrate on the CFL paradigm, with minimal attention given to the DFL paradigm. Furthermore, the current attack methodologies exhibit limitations, such as inconsistent efficacy and susceptibility to detection by defensive mechanisms. To address these research gaps, this paper presents the design and implementation of a novel attack strategy tailored for DFL. This proposed attack demonstrates broad effectiveness across various datasets, diverse ML model architectures, and multiple types of DFL overlay topologies.

\section{Problem Setup}
\subsection{DFL Training Process}
\hspace{1em}In the DFL framework, consider a network of \( K \) clients, each denoted as \( k \in \{1, 2, \dots, K\} \). In each communication round, each client \( k \) maintains and updates a local model \( w_t^k \) according to an aggregation algorithm \( A(\cdot) \) defined by this framework. The communication between clients can be represented as a graph \( G = (\mathcal{V}, \mathcal{E}) \), where \( \mathcal{V} \) is the set of nodes (clients) and \( \mathcal{E} \) is the set of edges (communication links) between the nodes. The neighborhood of a client \( k \), denoted as \( \mathcal{N}_k \subseteq \mathcal{V} \), contains all clients that are directly connected to \( k \) via an edge in \( \mathcal{E} \). Each client updates its model by training it locally and aggregating models from its neighbors. The network topology, represented by \( G \), plays a key role in determining how information is shared and how global knowledge is propagated through the network.

A general process of DFL can be divided into the following stages: First, each client \( k \) initializes its local model parameters \( w_0^k \). Then, each client trains its local model to further optimize these parameters. Gradient descent is generally used as the core optimization strategy during local training. Specifically, client \( k \) performs several epochs of local training using its private dataset \( P_k \) to minimize its local loss function \( \mathcal{L}(w, P_k) \) as defined in  Equation \ref{eq:weight_update}.

\begin{equation}
w_{t}^k \leftarrow w_t^k - \eta \nabla \mathcal{L}(w_t^k, P_k)
\label{eq:weight_update}
\end{equation}

where the gradient \( \nabla \ell(w_t^k, P_k) \) reflects the change in the loss function \( \ell \) with respect to the model parameters \( w_t^k \). The learning rate \( \eta \) controls the step size of the parameter updates at each iteration. By iteratively reducing the loss function, gradient descent effectively guides the model parameters toward a more optimal direction.

After each client \( k \) completes the training of its local model, it interacts with its neighbors \( \mathcal{N}_k \) and transmits the parameters of the current round of local model training to them, as defined by the network topology \( G \). Simultaneously, client \( k \) also receives model updates \( w_t^j \) from its neighbors \( j \in \mathcal{N}_k \) and follows an aggregation algorithm \( A(\cdot) \) to form a new model \( w_{t+1}^k \). The aggeration process can be defined in  Equation \ref{eq:aggreagation_update}.

\begin{equation}
w_{t+1}^k = A\left(w_t^k, \{w_t^j : j \in \mathcal{N}_k\}\right)
\label{eq:aggreagation_update}
\end{equation}

By continuously repeating the above processes of local model training and aggregation until the predefined number of rounds is completed, DFL effectively achieves a distributed and collaborative model optimization across the entire network, enhancing both the robustness and privacy of the learning process.

\begin{figure}[htbp]
    \centering
    \includegraphics[width=\columnwidth]{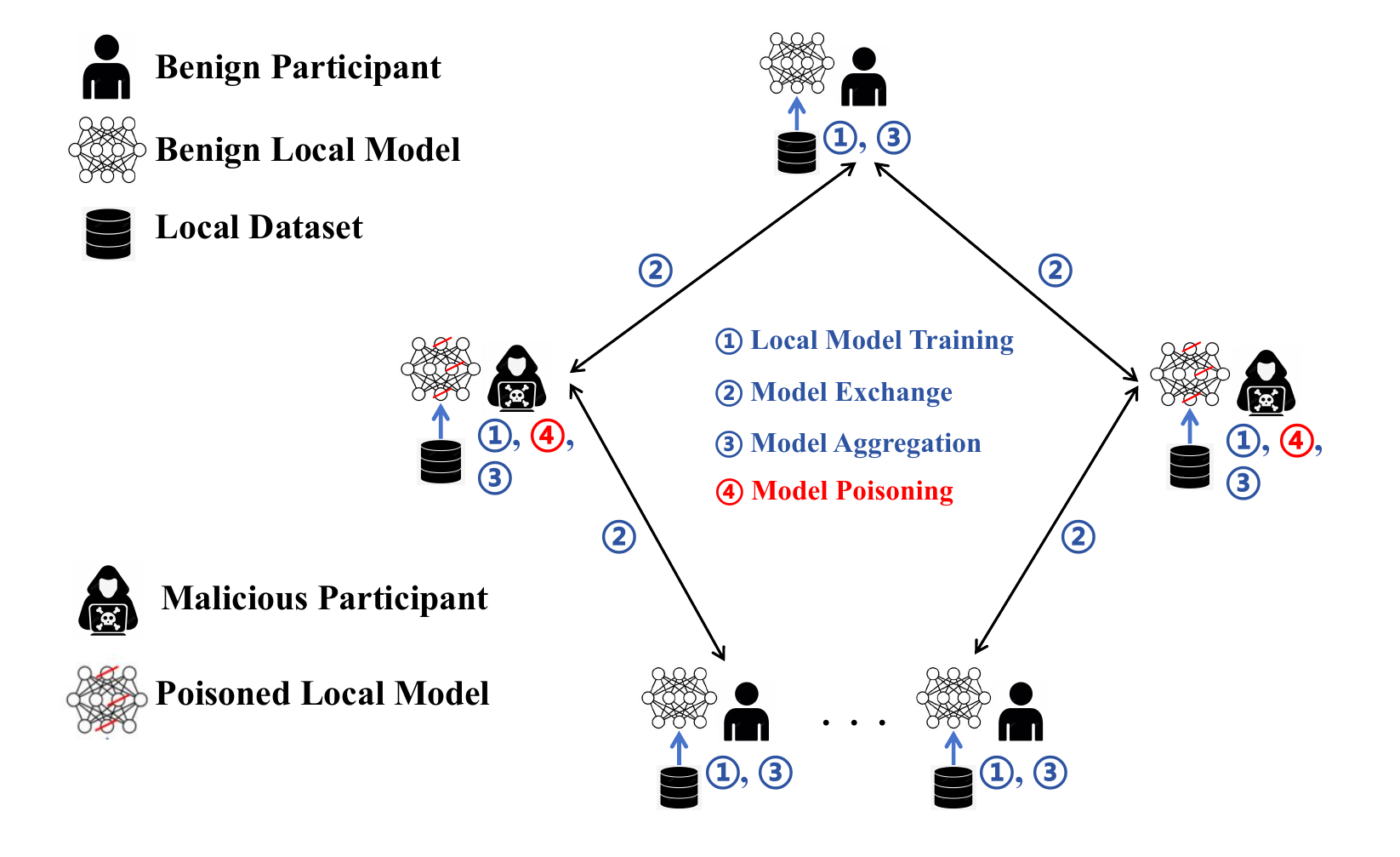} 
    \caption{DFL Process with Malicious Participants}
    \label{fig:figure2}
\end{figure}

Model poisoning attacks typically cause significant degradation in global model performance by maliciously manipulating model parameters. In the DFL framework, model poisoning attacks typically occur after the client has completed local model training. A malicious client deliberately tampers with its model parameters after training is complete and then sends these tainted parameters to its neighbors and participates in the model aggregation process. In this way, malicious clients are able to gradually contaminate the models of the entire federated learning system, ultimately leading to the degradation of the performance of the global model. The process of the DFL framework with malicious participants is illustrated in Figure\ref{fig:figure2}.

\subsection{Threat Model}
\hspace{1em}\textbf{Adversary Objective.}
The primary goal of the attacker is to minimize the performance of the global model in the DFL system by manipulating the updates to the client models under its control. Specifically, the attacker hopes to mislead the learning process of the global model by introducing carefully crafted malicious updates to the model parameters, ultimately leading to significant degradation in model accuracy or deviation in behavior.

\textbf{Adversary Capability.} 
The attacker has the ability to control a certain percentage of clients and has full access to and modify the local model updates of these clients. The attacker is able to send maliciously modified model parameters to neighboring nodes and participate in model aggregation after each round of training. In addition, the attacker can continuously observe and adjust the attack strategy during the training process to improve the stealth and effectiveness of the attack.

\textbf{Adversary Knowledge.} 
The attacker only has model information of all malicious participants and has no knowledge of the internal information of other benign clients. This knowledge level assumption is very strict, but it is consistent with the conditions of a realistic attack, which reflects the harmfulness of the designed model poisoning attack discussed in next section.

\section{DMPA Approach}
In this section, the general framework for applying the DMPA method in DFL is described, followed by a discussion of the research issues concerning Model Poisoning Attacks in DFL. The section then presents solutions to the problems outlined in the first section.
\subsection{Method Overview}

\begin{figure}[htbp]
    \centering
    \includegraphics[width=\columnwidth]{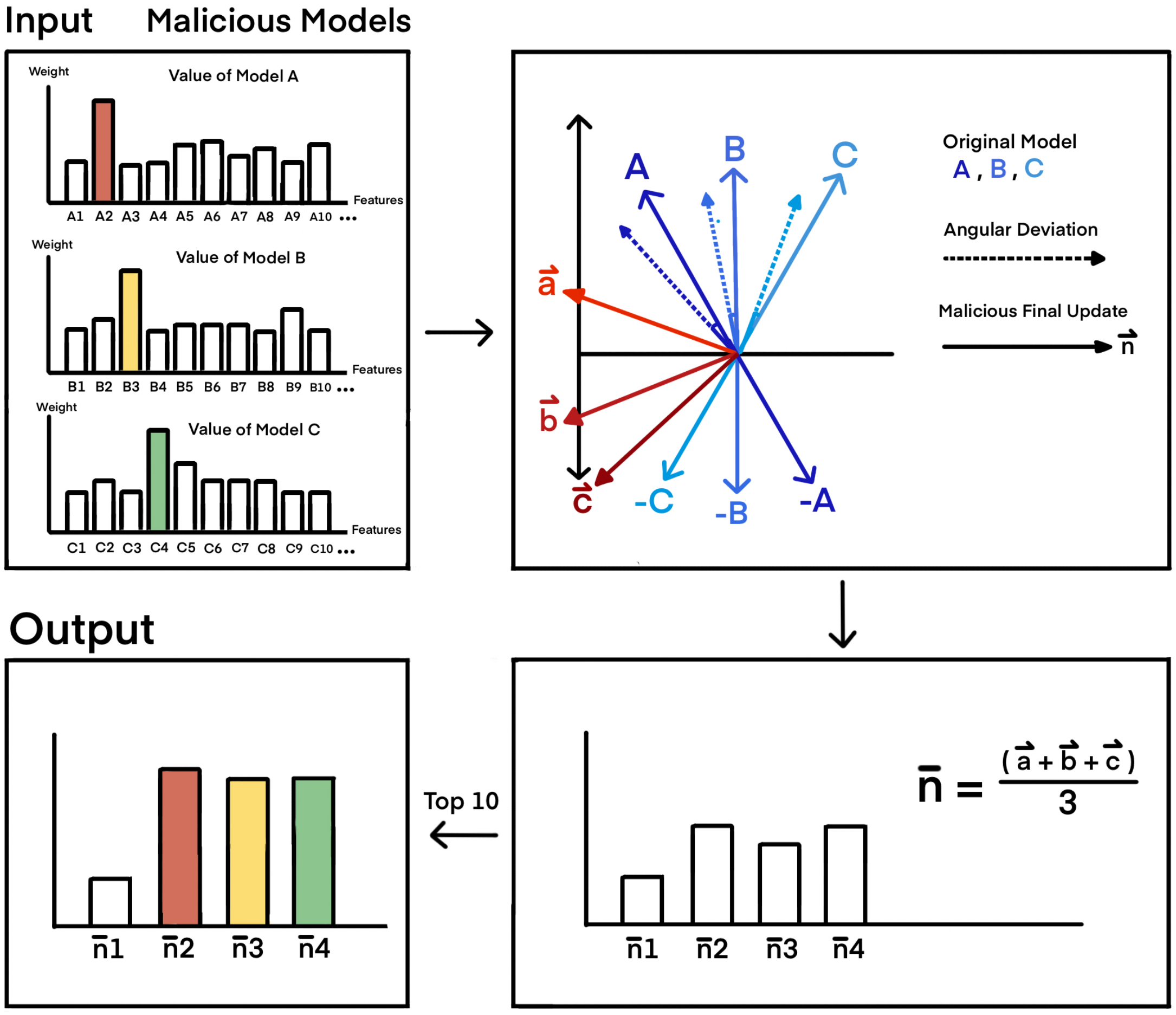} 
    \caption{Overview of DMPA Attack Process}
    \label{fig:figureDMPA}
\end{figure}

This workd proposes DMPA, an advanced model poisoning attack specifically designed to target DFL models. As illustrated in Figure \ref{fig:figureDMPA}, DMPA employs a tripartite attack strategy. In contrast to existing model poisoning attacks in FL (e.g., \cite{shejwalkar2021manipulating, baruch2019little, zhang2023denial}), which determine the attack direction based on model similarity, DMPA leverages the maximum eigenvalue of the correlation matrix to identify the optimal poisoning direction. The primary objective is to maximize the training loss of benign models after model aggregation.

As illustrated in Figure \ref{fig:figureDMPA}, DMPA is conducted the attack prior to the exchange of models among interconnected clients. There are three steps for DMPA to achieve the attack: (1) Find the correlation matrix according to the difference of the model parameter matrix, and calculate the eigenvectors of the maximum eigenvalue of the correlation matrix. Use this eigenvector to project on the model parameters to obtain the angular deviation; (2) Add this deviation with the inverse model parameters and the average adjusted model parameters; (3) Extract the uneven values from the average vector and use them to fill the corresponding positions in the average vector, thus maximizing the attack effect. Averaging the posterior vector and filling the original modified vector can prevent the attack effect from decreasing after the model is averaged, and malicious users upload malicious parameters to their connected benign clients for aggregation.

\begin{algorithm}[tb]
\caption{DMPA Attack Strategy}
\label{alg:algorithmDMPA}
\textbf{Input}: A matrix $\mathbf{U} \in \mathbb{R}^{d \times n}$ representing all updates, where each column $\mathbf{u}_i$ is a different update vector. \\
\textbf{Output}: The modified updates matrix $\mathbf{\mu}_{\text{new}}$. \\
\begin{algorithmic}[1] 
\STATE Compute the mean vector $\mathbf{\mu} = \frac{1}{n} \sum_{j=1}^{n} \mathbf{u}_{i,j,:}$
\STATE Center the updates: $\mathbf{V} = \mathbf{U} - \mathbf{\mu}$
\STATE Compute the covariance matrix: $\mathbf{C} = \frac{1}{n-1} \mathbf{V} \mathbf{V}^\top$
\STATE Compute the standard deviation: $\mathbf{T} = \sqrt{\text{diag}(\mathbf{C})}$
\STATE Compute the correlation matrix: $\mathbf{Y} = \frac{\mathbf{C}}{\mathbf{T} \mathbf{T}^\top}$ \\
\hspace{1em} \textit{where} $\oslash$ \textit{denotes element-wise division}
\STATE Compute eigenvalues and eigenvectors of $\mathbf{Y}$: $(\lambda, \mathbf{V}) = \text{eig}(\mathbf{Y})$
\STATE Find the principal eigenvalue: $\lambda_{\max} = \max(\lambda)$ and corresponding eigenvector $\mathbf{y}_{\max}$
\STATE Compute the projection: $\mathbf{P} = (\mathbf{y}_{\max}^\top \mathbf{U}) \cdot \mathbf{y}_{\max}$
\STATE Modify the updates: $\mathbf{U}^{\text{new}}  = -\mathbf{U} + \mathbf{P}$
\STATE Compute the new mean vector: $\mathbf{\mu}^{\text{new}} = \frac{1}{d} \sum_{j=1}^{d} \mathbf{U}_{j,:}$

\FOR{$i = 1$ to $d$}
    \STATE Compute mask $\mathbf{m}_i = \text{select\_top\_k\_params}(\mathbf{u}_i^2, 10)$
    \STATE Update mean vector: $\mathbf{\mu}^{\text{new}} = \mathbf{m}_i \odot \mathbf{U}_i^{\text{new}} + (1 - \mathbf{m}_i) \odot \mathbf{\mu}^{\text{new}}$
\ENDFOR
\STATE \textbf{return} $\mathbf{\mu}^{\text{new}}$

\STATE \textbf{Function} $\text{select\_top\_k\_params}(\mathbf{vector}, k\%)$:
\STATE \hspace{1em}$\text{sorted\_indices} = \text{argsort}(\mathbf{vector}, \text{descending=True})$
\STATE \hspace{1em} Compute $k = \left\lfloor \frac{\text{length of } \text{sorted\_indices} \times k\%}{100} \right\rfloor$
\STATE \hspace{1em} Initialize mask $\mathbf{m} = \mathbf{0}$ (same shape as $\mathbf{vector}$)
\STATE \hspace{1em} Set top $k$ indices: $\mathbf{m}[\text{sorted\_indices}_{:k}] = 1$
\STATE \hspace{1em} \textbf{return} $\mathbf{m}$
\end{algorithmic}
\end{algorithm}

\subsection{Attack Strategy}
Algorithm 1 provides the attack pseudocode of DMPA,  which explains how malicious clients execute the DPMA attack in DFL. This work defines the model of the received malicious client as $\mathbf{U}$. All malicious client models are transformed into column vectors, represented as $\mathbf{U} = [u_1, u_2, \ldots]$. During the attack phase, a decentralized approach is employed, leading to the computation of a new deviation, denoted as $v_{ij}$, using Equation \ref{eq:pingjun}. These resulting relative offsets collectively constitute the matrix $\mathbf{V}$.

\begin{equation}
    v_{ij} = u_{ij} - \bar{u}_j
    \label{eq:pingjun}
\end{equation}
where $\bar{u}_j$ is the average of the $j$ th model parameter.

To ensure the rigor of this analysis, it is essential to compute the overall standard deviation and provide an unbiased estimate. This necessitates the derivation of a correlation matrix. As demonstrated in Equation \ref{eq:duli}, the correlation matrix $\mathbf{V}$ is determined using the matrix $\mathbf{C}$.

\begin{equation}
    \mathbf{C} = \frac{1}{n-1} \mathbf{V}^T \mathbf{V}
    \label{eq:duli}
\end{equation}

The matrix $\mathbf{C}$ can be regarded as a covariance matrix, in which each element represents the correlation between subscript corresponding vector groups. However, in this paper, when finding out the weight of each model for gradient rise, it is necessary to get the weight of each model for gradient rise under independent assumptions. Therefore, the diagonal elements of the matrix $\mathbf{C}$ are extracted and the square root is found. After dimension reduction, the value can reflect the independent weight vector of each model $\mathbf{T}$.

\begin{equation}
\mathbf{Y} = \frac{\mathbf{C}}{\mathbf{T} \mathbf{T}^\top}
\label{eq:zbtouying}
\end{equation}

To obtain values characterized by independent weight and correlation deviation, Equation \ref{eq:zbtouying} computes the outer product $\mathbf{Y}$ by multiplying the correlation matrix with the independent weight vector. Here, $\mathbf{Y}$ is derived through a linear combination of correlation deviation and independent weight. This vector provides a more accurate representation of the system weight for each parameter column.

\begin{equation}
\mathbf{P} = \left( \mathbf{y}_{\max}^T \mathbf{U} \right)  \cdot \mathbf{y}_{\max}
\label{eq:touying}
\end{equation}

\begin{equation}
\mathbf{\mu}^{\text{new}} = - \mathbf{U} +\mathbf{P}
\label{eq:output}
\end{equation}

Upon obtaining the independent weight, the eigenvector $\mathbf{y}_{\max}$ associated with the maximum eigenvalue of the weight is computed. This eigenvector is then projected onto the original vector to determine the angular deviation, as described by Equation \ref{eq:touying}. Subsequently, the original model parameters are adjusted by incorporating the angular deviation, resulting in the final update, as delineated by Equation \ref{eq:output}.

\section{Evaluation}
\subsection{Experimental Setup}

\textbf{Datasets.} The experiments use CIFAR-10 \cite{krizhevsky2009learning}, MNIST \cite{deng2012mnist} and Fashion-MNIST \cite{xiao2017fashion} as validation datasets, which are widely used in validating model performance. The experiments adopt an independent identically distributed (IID) data partitioning method, which ensures that the data have the same statistical properties. The $\alpha$ parameter is defaulted to 100 as set in (\cite{shejwalkar2021manipulating}, \cite{tan2023collusive}, \cite{li2024fedimp}).

\textbf{Machine learning models.} Three different model architectures are chosen to correspond to different datasets. The batch size is set to 64 for all clients, and the random seed for each client is set to its corresponding ID value. A simple convolutional neural network (CNN) model with Conv2d, BatchNorm2d, 5 Depthwise Conv2d, and fully connected (linear) layers is used for the CIFAR-10 dataset, a model with three fully connected (linear) layers is used for the MNIST dataset, and a simple CNN model with 2 Conv2d and 2 fully connected layers is used for the Fashion-MNIST dataset.

\textbf{Measurement metrics.} The model F1 score is used to assess the attack performance. A lower F1 score indicates a better attack method. All results are the average of the F1 scores of all benign client models after last round of training, which enables a clear assessment of the impact of the MPA attack on the clients in the whole DFL network.

\textbf{Baseline defenses and attacks.} Three of the most effective model poisoning attacks in FL are chosen to attack DFL, namely Lie \cite{baruch2019little}, Min-Sum and Min-Max \cite{shejwalkar2021manipulating}. In DFL, the modifications of Lie, Min-Max and Min-Sum assume that the malicious clients are colluding, and thus these models can be shared among malicious clients \cite{li2023plato}. In addition, four defence methods are chosen; FedAvg, Krum, Trimmed Mean and Median. it is assumed that the malicious client has no knowledge of the benign model and only knows the models of all malicious clients in the current round

\begin{itemize}
    \item \textbf{Lie}: Updates the weights by subtracting the product of the standard deviation of the malicious parameters and the calculated perturbation range from the mean weights\cite{baruch2019little}.
    \item \textbf{Min-Max}: Computes the malicious gradient such that its maximum distance from any other gradient is upper bounded by the maximum distance between any two benign gradients\cite{shejwalkar2021manipulating}.
    \item \textbf{Min-Sum}: Ensures that the sum of squared distances between the malicious gradient and all benign gradients is upper bounded by the sum of squared distances between any benign gradient and the other benign gradients\cite{shejwalkar2021manipulating}.
\end{itemize}

\begin{table*}[ht]
\centering
\caption{Average F1 Score of Benign Clients in DFL with a Configuration Consisting of 40\% Malicious Clients for the MNIST, Fashion-MNIST, and CIFAR-10 Datasets in Fully-Connected, Ring, and Star Topologies.}
\resizebox{\textwidth}{!}{
\begin{tabular}{|c|c|cccc|cccc|cccc|}
\hline
\multirow{3}{*}{Dataset}      & \multirow{3}{*}{Aggregation Rule} & \multicolumn{4}{c|}{Fully}                                            & \multicolumn{4}{c|}{Ring}                                             & \multicolumn{4}{c|}{Star}                                    \\ \cline{3-14} 
                              &                                   & LIE               & Min-Max           & Min-Sum  & DMPA              & LIE               & Min-Max           & Min-Sum  & DMPA              & LIE               & Min-Max  & Min-Sum  & DMPA              \\ \hline
\multirow{4}{*}{MNIST}        & Median                            & 0.919523          & \textbf{0.828144} & 0.861094 & 0.847919          & 0.864083          & 0.727043          & 0.820756 & \textbf{0.691755} & 0.820844           & 0.821185 & 0.830490  & \textbf{0.820525} \\ \cline{2-14} 
                              & Trimmed mean                       & 0.827509          & 0.825695          & 0.843681 & \textbf{0.801535} & 0.864083          & 0.727043          & 0.820756 & \textbf{0.691755} & 0.822594          & 0.819780  & 0.825661 & \textbf{0.813209} \\ \cline{2-14} 
                              & Krum                              & 0.874433          & 0.811460           & 0.885670  & \textbf{0.013312} & 0.873979          & 0.867974          & 0.840262 & \textbf{0.740244} & 0.819409          & 0.812599 & 0.822512 & \textbf{0.582298} \\ \cline{2-14} 
                              & Fed\_avg                          & 0.825292          & 0.826278          & 0.834523 & \textbf{0.802228} & 0.833553          & \textbf{0.64567}  & 0.827210  & 0.672756          & 0.821639          & 0.820869 & 0.825211 & \textbf{0.815387} \\ \hline
\multirow{4}{*}{CIFAR-10}    & Median                            & 0.410982          & 0.679403          & 0.706879 & \textbf{0.044529} & \textbf{0.280601} & 0.421047          & 0.446616 & 0.314470           & 0.548596          & 0.665249 & 0.652866 & \textbf{0.490103} \\ \cline{2-14} 
                              & Trimmed mean                       & 0.652813          & 0.734104          & 0.732013 & \textbf{0.016994} & \textbf{0.258510} & 0.400051          & 0.395156 & 0.305964          & 0.596924          & 0.636560  & 0.658856 & \textbf{0.193741} \\ \cline{2-14} 
                              & Krum                              & \textbf{0.230500} & 0.625591          & 0.625531 & 0.569555          & \textbf{0.372232}   & 0.545091          & 0.646702 & 0.433777          & \textbf{0.538895} & 0.600382 & 0.594952 & 0.569471          \\ \cline{2-14} 
                              & Fed\_avg                          & 0.674934          & 0.740961          & 0.737501 & \textbf{0.016994} & 0.346825          & 0.440712          & 0.449766 & \textbf{0.076015} & 0.617962          & 0.640634 & 0.652659 & \textbf{0.213768} \\ \hline
\multirow{4}{*}{Fashion-MNIST} & Median                            & 0.887029          & 0.892688          & 0.898451 & \textbf{0.881820}  & 0.843861          & 0.800340           & 0.885512 & \textbf{0.773666} & 0.883854          & 0.878469 & 0.887852 & \textbf{0.878103} \\ \cline{2-14} 
                              & Trimmed mean                       & 0.893127          & 0.885061          & 0.903730  & \textbf{0.863626} & 0.837478          & 0.800340           & 0.885512 & \textbf{0.773666} & 0.887656          & 0.878575 & 0.898533 & \textbf{0.874800}   \\ \cline{2-14} 
                              & Krum                              & 0.875101          & 0.862932          & 0.896126 & \textbf{0.058431} & 0.806988           & \textbf{0.453659} & 0.879798 & 0.767491          & 0.860227          & 0.864078 & 0.884784 & \textbf{0.597608} \\ \cline{2-14} 
                              & Fed\_avg                          & 0.895660          & 0.884624          & 0.897920  & \textbf{0.861161} & 0.880633          & \textbf{0.754967} & 0.894226 & 0.769304          & 0.890041          & 0.873614 & 0.892970  & \textbf{0.876514} \\ \hline
\end{tabular}}

\end{table*}

\textbf{DFL overlay topologies.} Three different DFL overlay topologies are employed in the experiments to evaluate the compliance of the proposed attack DPMA in different types of federation. 
\begin{itemize}
    \item \textbf{Fully:} Each node is directly connected to every other node.
    \item \textbf{Star:} All nodes are connected through a central node.
    \item \textbf{Ring:} Each node is connected to two neighboring nodes, forming a closed loop.
\end{itemize}

\textbf{Training environment.} 
The experiments are conducted using Python 3 and PyTorch, running on an NVIDIA T4 GPU. A total of 10 rounds are run, with each client performing 3 local training epochs in each round. The experiments involve 10 clients, and the DFL experimental programs for all nodes are executed sequentially.

\subsection{Experimental Results}
\textbf{Compare DMPA with state-of-art model poisoning attacks.} In this section, the attack method proposed in this study is compared with the current state-of-the-art model poisoning attack methods, including Lie\cite{baruch2019little} and Min-Max and Min-Sum \cite{shejwalkar2021manipulating}. Three topologies (Fully, Star, and Ring) are used for the comparison experiments, and IID data is used for the experiments. The experimental results are based on a configuration of 40\% malicious clients and 60\% benign clients, and calculate the average F1 scores in the final round (smaller F1 scores represent more effective attacks). The experimental results are shown in Table I. Bolded data indicates the best results.

Firstly, from the experimental results, when the proportion of malicious participants is 40\%. The results show that DMPA has a significant attack effect in DFL with different topologies, different aggregation methods, and different datasets, and its F1 scores are all the lowest, indicating that DMPA has the best attack effect, which fully verifies the effectiveness of the method on attack.

\textbf{Impact of overlay network topology.} In fully connected networks, DMPA shows extreme attackability in several experiments. For example, in experiments on the MNIST dataset corresponding to the Krum aggregation method, the CIFAR-10 dataset corresponding to the Median, Trimmed Mean and FedAvg aggregation methods, and the Fashion-MNIST dataset corresponding to the Krum aggregation method, DMPA achieves a score of less than 0.1. This shows that DMPA effectively influences the gradient descent process of 60\% benign clients and successfully fills the difference in gradient descent by increasing the loss. This shows that DMPA has been validated for its reasonable design, which can invalidate the gradient descent by increasing and filling in the gradient difference. In contrast, although the other three attack methods achieve the attack by influencing the direction and magnitude of the model update, most of the average F1 scores among the 60\% benign clients are greater than 0.8, indicating that their attack effects are more limited.

The overall F1 score of DMPA in the ring network is low and close to the lowest of the other attack methods. The F1 scores of DMPA are all less than 0.8, showing high stability. This may be due to the fact that in ring networks, 60\% of the benign clients are buffered to some extent due to the longer chain, mitigating the impact of the attack, which is a major advantage of DFL. As for the other methods, half of them have F1 scores above 0.8, indicating that these methods are not stable enough for model poisoning attacks in the ring structure and are prone to only changing the direction of the model and not effectively affecting other benign clients. This further demonstrates that DMPA exhibits stable and effective attacks that can generally affect benign clients in the DFL ring network.
In the Star network, DMPA also shows its advantages, further proving the effectiveness of the method in this paper. Since the results of CIFAR-10 are the most representative for studying the attack ratio among the three datasets, the CIFAR-10 dataset will be selected in the following to further investigate the impact of the malicious client ratio on the overall experimental results.

\begin{figure}[htbp]
    \centering
    \begin{subfigure}[b]{\columnwidth}
        \centering
        \includegraphics[width=\textwidth]{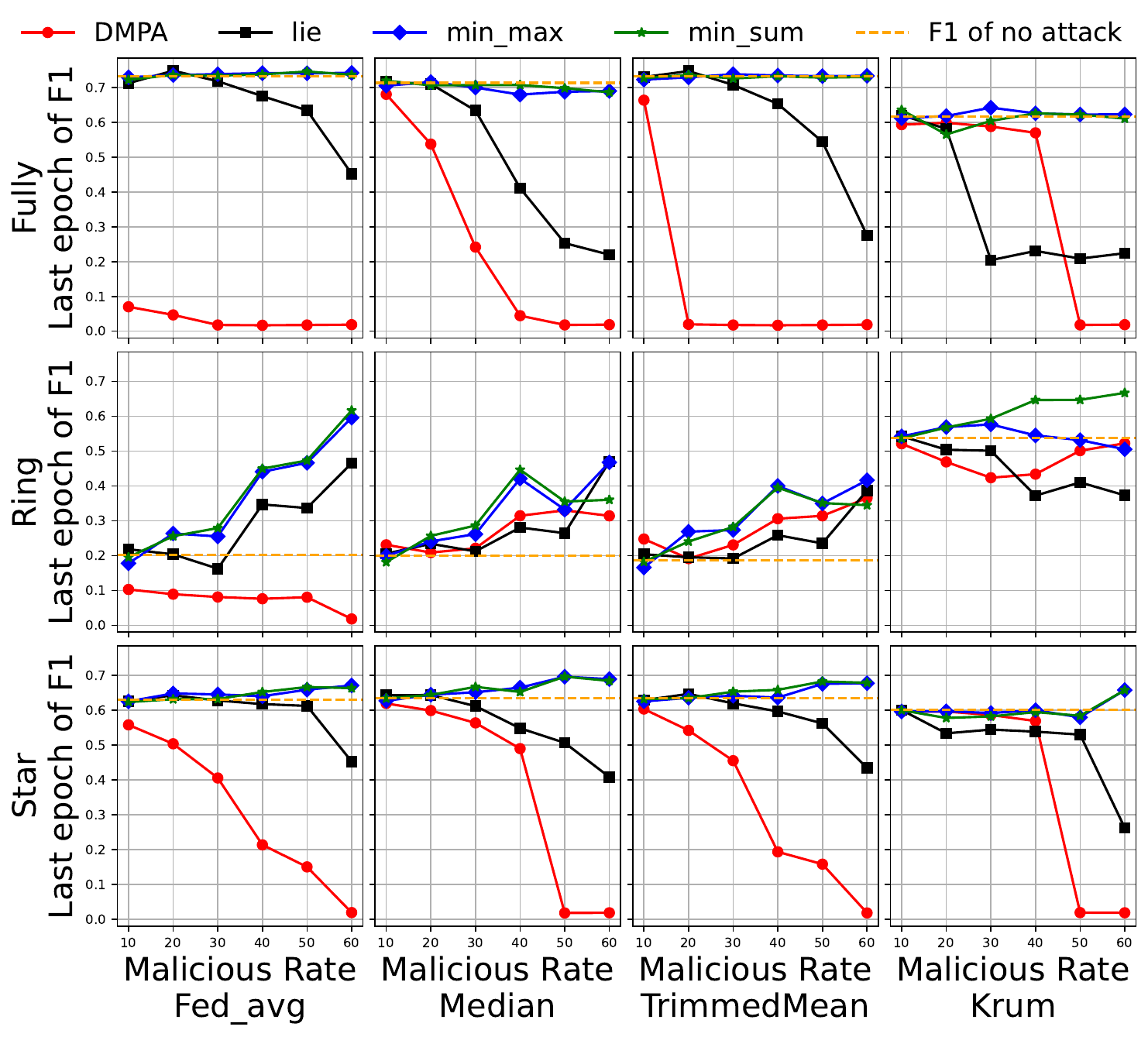}
        \caption{CIFAR-10 Dataset Attack Performance}
        \label{fig:cifar-attack}
    \end{subfigure}

    \begin{subfigure}[b]{\columnwidth}
        \centering
        \includegraphics[width=\textwidth]{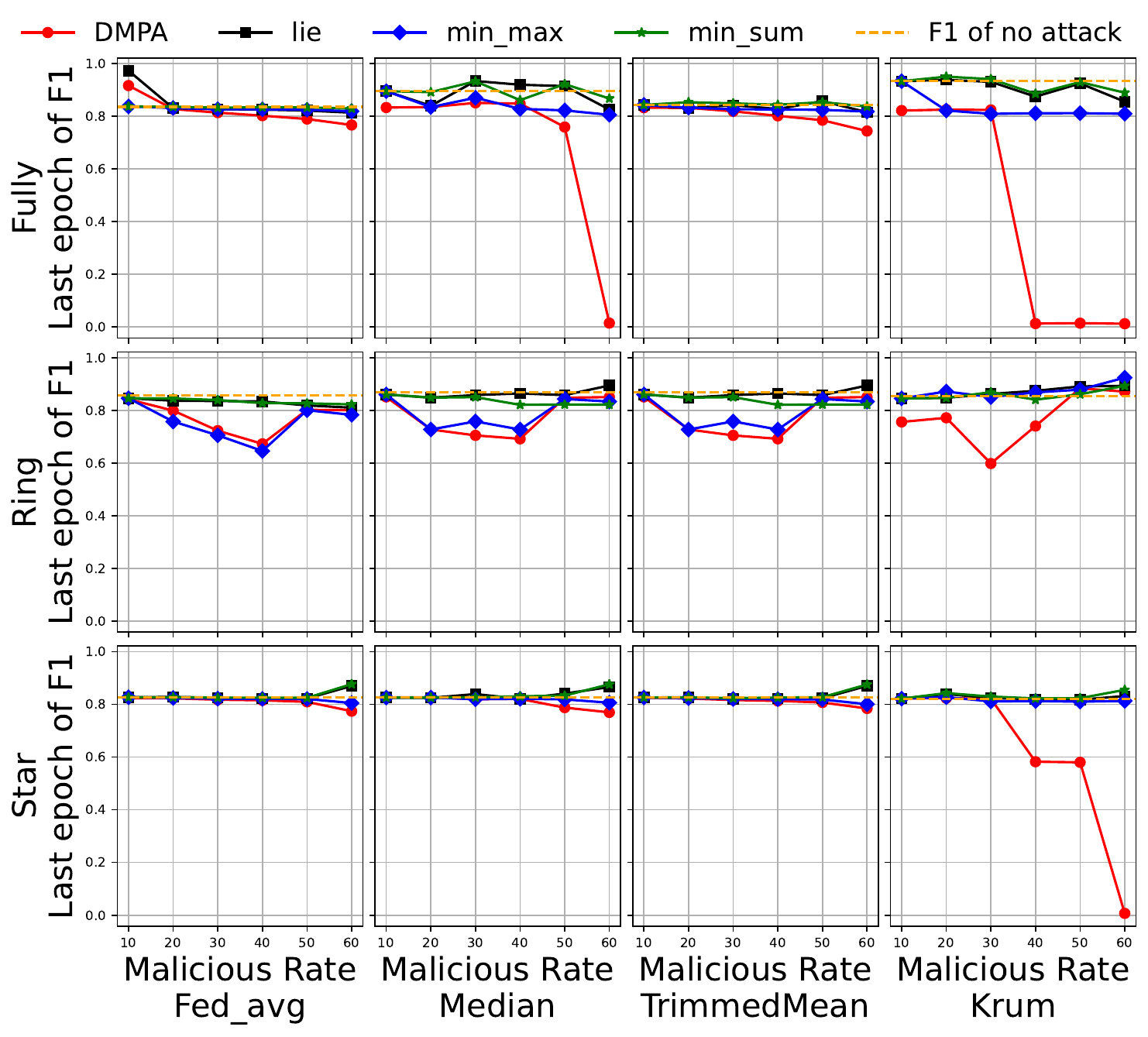}
        \caption{MNIST Dataset Attack Performance}
        \label{fig:mnist-atack}
    \end{subfigure}

    \begin{subfigure}[b]{\columnwidth}
        \centering
        \includegraphics[width=\textwidth]{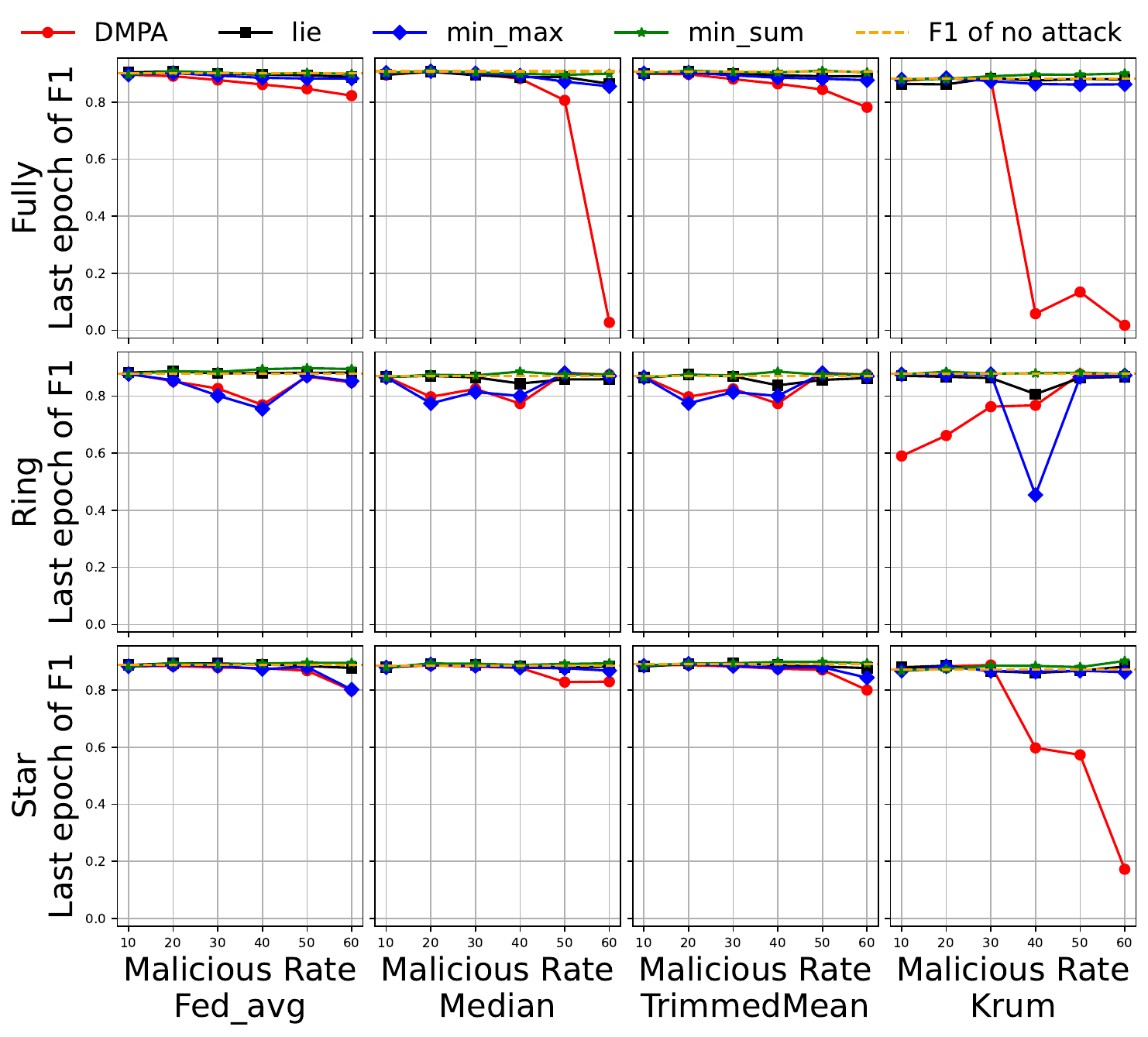}
        \caption{Fashion-MNIST Dataset Attack Performance}
        \label{fig:fmnist-attack}
    \end{subfigure}
    \caption{Attack Performance in CIFAR-10, MNIST, and Fashion-MNIST with Different Malicious Clients Rate}
\end{figure}

\textbf{Impact of increasing the percentage of malicious Participants.} Figure \ref{fig:cifar-attack} to Figure \ref{fig:fmnist-attack} shows the impact of MPA on the F1 score of benign clients after the last round of aggregation in datasets, as the rate of malicious nodes increases from 10\% to 60\% in the DFL environment. The orange of the dotted line is no attack in each aggregation. 

As can be seen, the DMPA method shows the best attack effect in many results to achieve effective attacks. Compared with other methods on different data sets, DMPA has the influence of attacks. However, in different datasets, this paper finds that the more complex the network is, the more effective its influence is. When more network layers are used (for example, CIFAR-10 uses more complex networks), its attack effectiveness is higher than other methods. Obviously, the higher the complexity of the network, the more influence it has. Overall, the method in this paper is more effective than other attack methods at present.

\section{Conclusion and Future Work}

This study proposes DMPA, a general framework designed to execute model poisoning attacks within DFL systems. The DMPA framework leverages feature angle deviations to ascertain the most effective attack strategy. In contrast to prior attack methodologies that rely on imprecise or heuristic approaches, DMPA employs the model's numerical properties for its computations, thereby maintaining the model's numerical integrity across iterations. Empirical results indicate that DMPA achieves superior attack efficacy compared to existing state-of-the-art model poisoning techniques, underscoring its substantial practical relevance.

Future research is planned to explore both offensive and defensive dimensions. From an offensive standpoint, existing studies predominantly utilize data distributed in an IID manner, thereby simplifying the attack process. However, the complexity of attacks escalates when data distribution is non-IID. Consequently, future research will focus on devising strategies for effective assaults under non-IID conditions. From a defensive perspective, the proposed DPMA underscores the efficacy of using feature angle deviations as a potent attack vector, providing valuable insights for the development of robust mechanisms to protect against potential threats.

\end{document}